\documentclass{article}
\usepackage[final]{neurips_2022}
\usepackage{microtype, graphicx, subfigure, booktabs, multirow}
\usepackage[utf8]{inputenc}
\usepackage[T1]{fontenc}
\usepackage{hyperref, url, amsfonts, nicefrac, xcolor}
\usepackage{amsmath, amssymb, mathtools, amsthm, bbm, lipsum}
\usepackage[frak = euler]{mathalpha}
\usepackage[capitalize,noabbrev]{cleveref}
\usepackage{enumerate}
\usepackage[normalem]{ulem} 
\usepackage{pythonhighlight} 
\usepackage{hyperref} 

\newcommand{\model}[1]{\texttt{#1}}

\theoremstyle{plain}

\theoremstyle{definition}

\theoremstyle{remark}

\newcommand{\ours}{\model{{TimeGPT}}}
\newcommand{\ZeroModel}{\model{ZeroModel}}
\newcommand{\HistoricAverage}{\model{HistoricAverage}}

\newcommand{\SeasonalNaive}{\model{SeasonalNaive}}
\newcommand{\ThetaModel}{\model{Theta}}
\newcommand{\DOTheta}{\model{DOTheta}}
\newcommand{\ETS}{\model{ETS}}
\newcommand{\CES}{\model{CES}}
\newcommand{\ADIDA}{\model{ADIDA}}
\newcommand{\IMAPA}{\model{IMAPA}}
\newcommand{\CrostonClassic}{\model{CrostonClassic}}
\newcommand{\LGBM}{\model{LGBM}}
\newcommand{\DeepAR}{\model{DeepAR}}
\newcommand{\TFT}{\model{TFT}}
\newcommand{\LSTM}{\model{LSTM}}
\newcommand{\NHITS}{\model{NHITS}}

\title{TimeGPT-1}
\author{%
  Azul Garza\textsuperscript{*}, Cristian Challu\textsuperscript{*}, Max Mergenthaler-Canseco \thanks{Authors contributed equally.}\\
  Nixtla\\
  San Francisco, CA, USA \\
  \texttt{\{azul,cristian,max\}@nixtla.io}
}

\begin{document}

\maketitle

\begin{abstract}

In this paper, we introduce \ours, the first foundation model for time series, capable of generating accurate predictions for diverse datasets not seen during training. We evaluate our pre-trained model against established statistical, machine learning, and deep learning methods, demonstrating that \ours\ zero-shot inference excels in performance, efficiency, and simplicity. Our study provides compelling evidence that insights from other domains of artificial intelligence can be effectively applied to time series analysis. We conclude that large-scale time series models offer an exciting opportunity to democratize access to precise predictions and reduce uncertainty by leveraging the capabilities of contemporary advancements in deep learning.

\end{abstract}

\section{Introduction}\label{section:introduction}

Uncertainty is an intrinsic aspect of life, a constant element that humans have tirelessly sought to navigate and comprehend. From the traditions established by ancient civilizations to the sophisticated research endeavors in our contemporary world, brilliant minds have ceaselessly strived to anticipate the distribution of possible future events, crafting systematic approaches to unveil the prospective future.

The aspiration to predict potential outcomes, foundational across a multitude of disciplines, reflects a deep-seated human tendency to anticipate, strategize, and mitigate risks. The goal to reduce uncertainty about what will come next maps to numerous real-world applications: from understanding economic cycles and trends to discerning consumer consumption patterns; from optimizing electricity demand for energy production and grid management to aligning capacity and infrastructure for servers, workers, and machines.

Time series—data ordered chronologically—constitutes the underlying fabric of systems, enterprises, and institutions. Its impact spans from measuring ocean tides to tracking the daily closing value of the Dow Jones. This type of data representation is indispensable in sectors such as finance, healthcare, meteorology, social sciences, and others, where discerning temporal patterns, trends, and cyclical variations is crucial for forecasting future values and informing decision-making processes.

However, the current theoretical and practical understanding of time series hasn't yet achieved a level of consensus among practitioners that mirrors the widespread acclaim for generative models in other fundamental domains of the human condition, like language and perception. Our field is still divided in their assessment of the efficacy of deep learning for forecasting tasks. Efforts in forecasting science have fallen short of fulfilling the promises of genuinely universal pre-trained models.

In this paper, we embark on a novel path and introduce \ours, the first pre-trained foundation model for time series forecasting that can produce accurate predictions across a diverse array of domains and applications without additional training. A general pre-trained model constitutes a groundbreaking innovation that opens the path to a new paradigm for the forecasting practice that is more accessible and accurate, less time-consuming, and drastically reduces computational complexity.

\section{Background}\label{section:background}

Regarding the superiority of deep learning approaches, the forecasting community is currently divided. A unified approach has yet to be established. Recently, these diverging paradigms have increasingly challenged each other, questioning the usefulness, accuracy, and complexity of new developments. Despite the success of deep learning architectures in other fields, some time series practitioners have demonstrated that some proposed innovations in the field don't fulfill their claims or expectations. \footnote{It must be noted, that although this characterization fails to fully account for specific cases of hybrid forecasting the main claims remain valid. For further discussion see: \citep{smyl2020hybrid} and \citep{januschowski2020criteria}}

Historically, statistical methods such as ARIMA, ETS, MSTL, Theta, and CES have been reliably employed across various domains. In the past decade, machine learning models like XGBoost and LightGBM have gained popularity, demonstrating promising results in both public competitions and practical applications.

However, with the advent of deep learning, a paradigm shift in time series analysis has occurred. Deep learning methods have become popular in academia and for large-scale industrial forecasting applications \citep{10.1145/3533382}.

Given their global approach, deep learning methods offer significant advantages over statistical local methods in terms of scalability, flexibility, and potential accuracy. Additionally, their ability to learn intricate data dependencies effectively bypasses the need for complex feature engineering necessary for other global methods like LightGBM or XGBoost. Consequently, deep learning-based time series models aim to simplify the forecasting pipeline and enhance scalability. Their ability to handle large volumes of data and capture long-term dependencies positions them advantageously for complex forecasting tasks in an era of ever-growing data volumes.

However, opinions among academic researchers and practitioners diverge regarding these promises. Various researchers and practitioners have challenged the basic assumption of increased accuracy, presenting evidence showing that simpler models outperform more sophisticated approaches; with less cost and complexity. Conversely, some industry leaders report that the deep learning approach has enhanced their results and simplified their analysis pipelines \citep{kunz2023deep}.

In the current historical context, where the superior capabilities of deep learning models are undeniable for natural language processing (NLP) and computer vision (CV), it's noteworthy that the time series analysis field remains skeptical of the performance of neural forecasting methods. 

We believe this skepticism arises from:

\begin{itemize}
    \item Misaligned or poorly defined evaluation settings: Unlike other fields that have benefited from the introduction of ideal testing datasets such as ImageNet for computer vision, the publicly available datasets for time series do not possess the necessary scale and volume for deep learning methods to excel.
    \item Suboptimal models: Given the limited and specific datasets, even well-conceived deep learning architectures might struggle with generalization or require considerable effort to find optimal settings and parameters.
\end{itemize} 

Furthermore, the lack of standardized large-scale datasets that cater to the requirements of deep learning methods could also be hindering progress in this area. While other fields have benefited from benchmark datasets and clear evaluation metrics, the time series community still needs to develop such resources to foster innovation and validate new techniques.\footnote{
For a detailed analysis of the state of our field, we refer the interested reader to notable systematization such as \citep{de200625} and \citep{10.1145/3533382, januschowski2020criteria}.}

In this paper, we demonstrate that larger and more diverse datasets enable more sophisticated models to perform better across various tasks. \ours\ is the first foundation model that consistently outperforms alternatives with minimal complexity. Further researching the improvements of foundation models for time series could potentially usher in a new chapter in the field, fostering a more profound understanding of temporal data and enhancing the accuracy and efficiency of forecasts.
\section{Literature Review}\label{section:literature}

\begin{figure*}[t]
\centering
\subfigure[Single series forecasting]{\label{fig:singleforecast}
\includegraphics[width=0.5\linewidth]{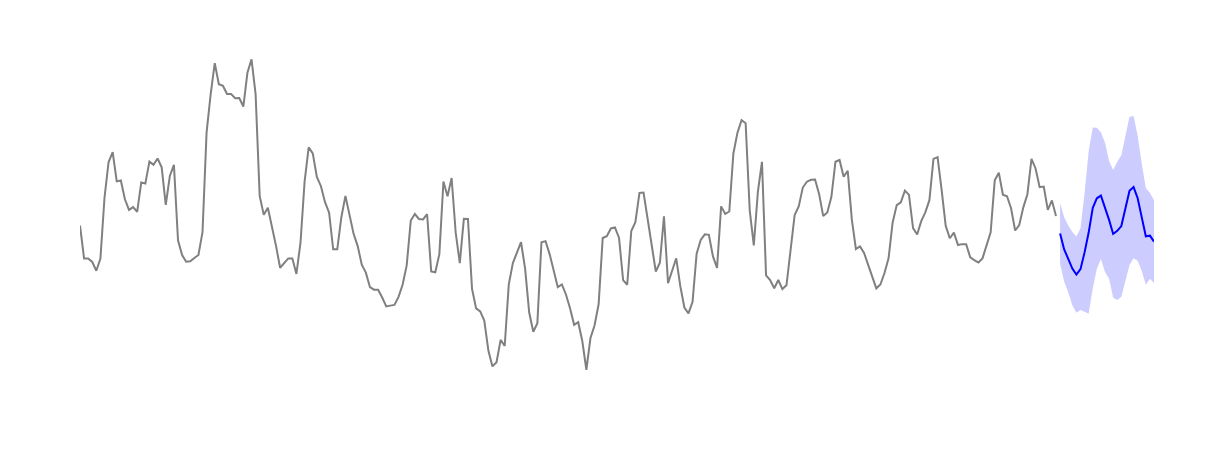}}
\subfigure[Multiple series forecasting]{\label{fig:multiforecast}
\includegraphics[width=0.4\linewidth]{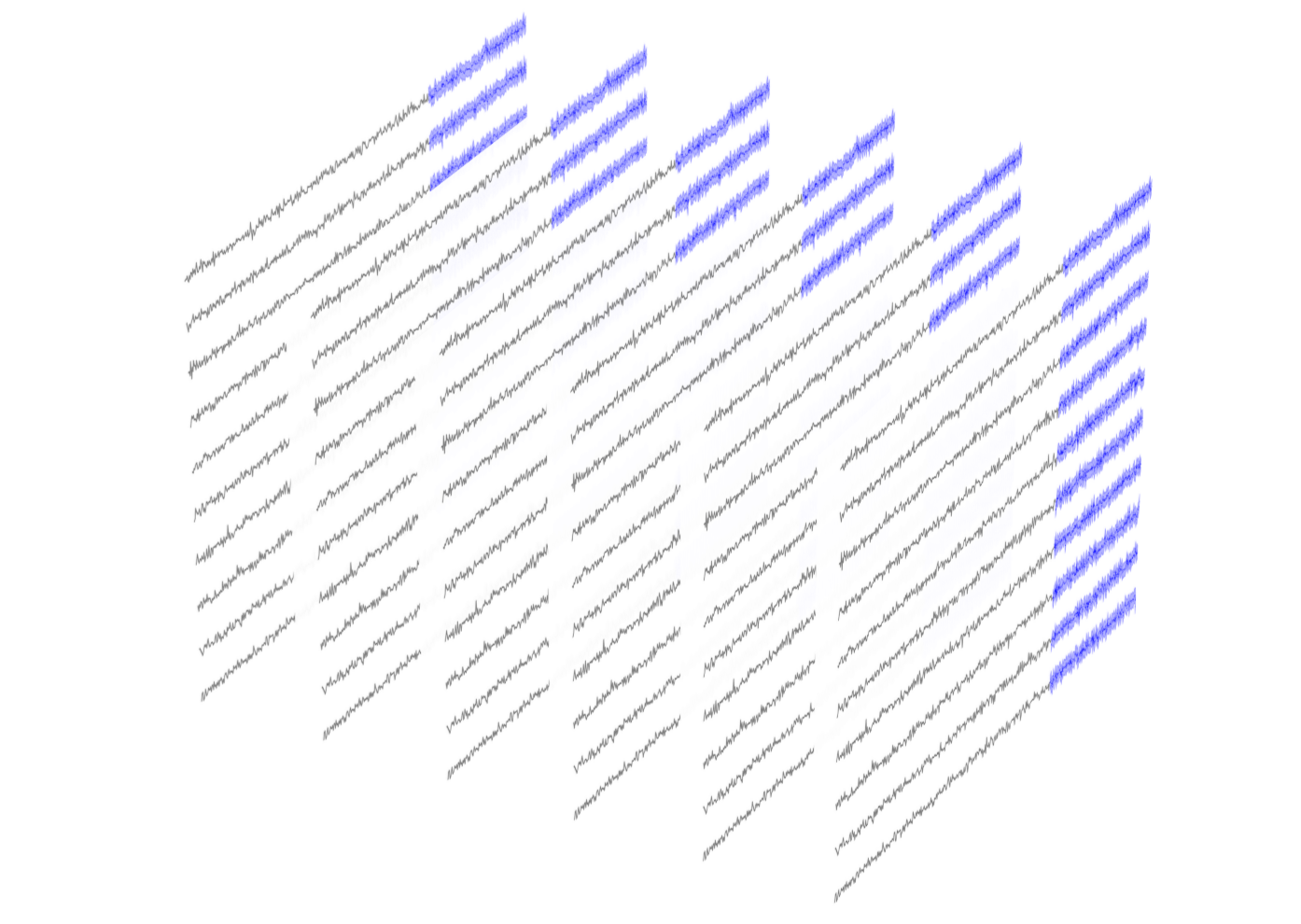}}
\caption{Illustration of single series forecasting and multiple series forecasting} 
\label{fig:illustration}
\end{figure*}

Deep Learning forecasting models have become a prominent area of research, driven by their success in recent famous competitions, including \citep{makridakis2020m4, makridakis2022m5}, and their applicability to large-scale tasks in the industry. \citep{10.1145/3533382} presents a comprehensive review and taxonomy of neural forecasting models and their applications.

Initial Deep Learning time series forecasting successes stemmed from the adaptation of established architectures, namely Recurrent Neural Networks (RNN) and Convolution Neural Networks (CNN), initially designed for natural language processing (NLP) and computer vision (CV), respectively. RNNs served as the backbone for popular models like DeepAR \citep{salinas2020deepar} for probabilistic forecasting and the ESRNN \citep{smyl2020hybrid}, winner of the M4 Competition. CNNs demonstrated superior performance than RNNs in multiple tasks on sequential data, as shown in \citep{bai2018empirical}. They now constitute a popular building block, as models like DPMN \citep{olivares2023probabilistic} and TimesNet \citep{wu2022timesnet} use. Feed-forward networks, due to their low computational costs and efficiency, are also frequently used, with notable examples including the N-BEATS \citep{oreshkin2019n, olivares2022neural} and NHITS \citep{challu2023nhits}.

Transformer-based models \citep{vaswani2017attention} are gaining popularity in recent years, as they are demonstrating remarkable performance in large-scale settings \citep{kunz2023deep} and complex tasks, such as long sequence forecasting. The earlier examples include the TFT \citep{lim2021temporal} and MQTransformer \citep{eisenach2020mqtransformer}, both with multi-quantile capabilities. The Informer introduced Transformers for long sequence forecasting through the Prob-sparse self-attention mechanism \citep{zhou2021informer}. This concept has since been further refined through various forms of inductive bias and attention mechanisms in models like the Autoformer \citep{wu2021autoformer}, FEDformer \citep{zhou2022fedformer}, and PatchTST \citep{nie2022time}.

The potential of foundation models, namely large-scale models pre-trained on a large dataset and later fine-tuned for specific tasks, remains relatively under-explored for time series forecasting tasks. There are, however, early indicators of the possibility of forecasting foundational models. For instance, \citep{oreshkin2021meta} showed that pre-trained models can be transferred between tasks without performance degradation. Additionally, \citep{kunz2023deep} provided evidence on the existence of scaling laws on data and model sizes for Transformer architectures on time series forecasting tasks.

\section{Foundation model for time series}\label{section:GeneralModels}

Foundation models rely on their capabilities to generalize across domains, particularly in new datasets that were not available during training. We understand, accordingly, transfer learning as the capacity to apply knowledge gleaned from one task to solve new tasks. Next, we explain the concept of transfer learning, building upon previous studies in time series forecasting \citep{oreshkin2021meta, olivares2023transferability}.



A forecasting model provides a function $f_\theta:\mathcal{X} \mapsto \mathcal{Y}$, with $\mathcal{X}$ the feature space and $\mathcal{Y}$ the dependent variable space. We consider the setting with $\mathcal{X}=\{\mathbf{y}_{[0:t]}, \mathbf{x}_{[0:t+h]}\}$ and $\mathcal{Y}=\{\mathbf{y}_{[t+1:t+h]}\}$, where $h$ is the forecast horizon, $\mathbf{y}$ is the target time series, and $\mathbf{x}$ are exogenous covariates. The forecasting task objective is to estimate the following conditional distribution:

\begin{equation}
    \mathbb{P}\left(\mathbf{y}_{[t+1:t+h]}|\; \mathbf{y}_{[0:t]}, \mathbf{x}_{[0:t+h]}\right) = f_\theta(\mathbf{y}_{[0:t]}, \mathbf{x}_{[0:t+h]})
\end{equation}

Transfer-learning refers to pre-training a model on a (usually large) source dataset $D_s=\{(\mathbf{X},\mathbf{y})|\; \mathbf{X} \in \mathcal{X},\; \mathbf{y} \in \mathcal{Y}\}$, to improve its performance on a new forecasting task with target dataset $D_t$. This paper considers two cases of transfer learning: zero-shot learning and fine-tuning. In the first case, the pre-trained model is directly transferred to solve the new forecasting task without re-training its parameters $\theta$ on the new dataset. Conversely, in fine-tuning, the model is further trained on the new dataset (starting from pre-trained parameters).

The core idea of the presented foundation model is to leverage these principles by training it on the largest publicly available time series dataset to date, leveraging scaling laws on the dataset and model sizes. A diverse dataset, in terms of breadth and depth, allows \ours\ to glean insights from an unprecedented array of temporal patterns across multiple domains.


\section{\ours}\label{section:methodology}

\subsection{Architecture}

\ours\ is a Transformer-based time series model with self-attention mechanisms based on \citep{vaswani2017attention}. \ours\ takes a window of historical values to produce the forecast, adding local positional encoding to enrich the input. The architecture consists of an encoder-decoder structure with multiple layers, each with residual connections and layer normalization. Finally, a linear layer maps the decoder's output to the forecasting window dimension. The general intuition is that attention-based mechanisms are able to capture the diversity of past events and correctly extrapolate potential future distributions.

\begin{figure*}[t] %
\centering
\includegraphics[width=\linewidth]{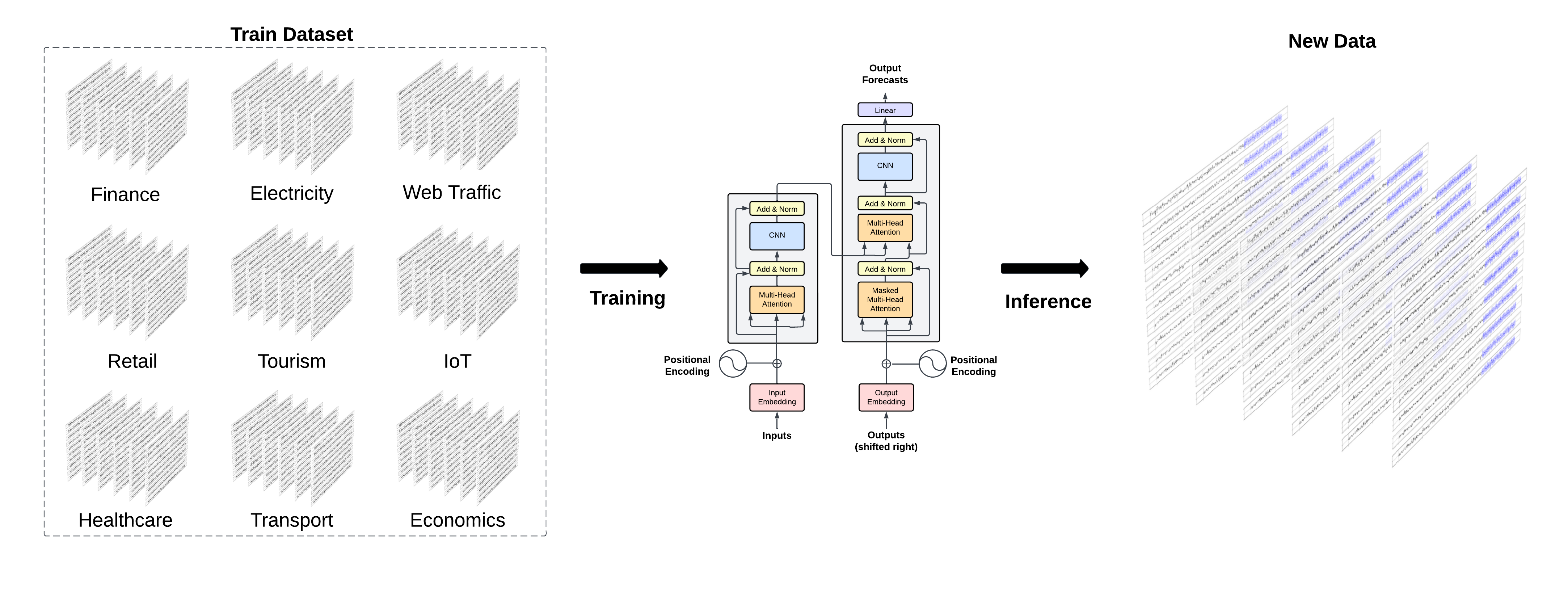}
\caption{\ours\ was trained in the largest collection of publicly available time series, and can forecast unseen time series without re-training its parameters.}
\label{fig:architecture}
\end{figure*}

The development of a generalized global model for time series entails numerous challenges, primarily due to the complex task of handling signals derived from a broad set of underlying processes. Characteristics such as frequency, sparsity, trend, seasonality, stationarity, and heteroscedasticity present distinct complications for both local and global models. Therefore, any foundational forecasting model must possess the ability to manage such heterogeneity. Our model, \ours, is engineered to process time series of varied frequencies and characteristics while accommodating different input sizes and forecasting horizons. This adaptability is largely attributable to the underlying transformer-based architecture upon which \ours\ is built.

It should be noted that \ours\ is not based on an existing large language model (LLM). While \ours\ follows the same principle of training a large transformer model on a vast dataset, its architecture is specialized in handling time series data and trained to minimize the forecasting error.

\subsection{Training dataset}

\ours\ was trained on, to our knowledge, the largest collection of publicly available time series, collectively encompassing over 100 billion data points. This training set incorporates time series from a broad array of domains, including finance, economics, demographics, healthcare, weather, IoT sensor data, energy, web traffic, sales, transport, and banking. Due to this diverse set of domains, the training dataset contains time series with a wide range of characteristics.

In terms of temporal patterns, the training dataset contains series with multiple number of seasonalities, cycles of different lengths, and various types of trends. In addition to the temporal patterns, the dataset also varies in terms of noise and outliers, offering a robust training environment. Some series contain clean, regular patterns, while others feature significant noise or unexpected events, providing a broad spectrum of scenarios for the model to learn from. Most of the time series were included in their raw form; the processing was limited to format standardization and filling in missing values to ensure data completeness.

The selection of such a diverse training set is critical for developing a robust foundational model. This diversity encompasses the complex realities of non-stationary real-world data, where trends and patterns can shift over time due to a multitude of factors. Training \ours\ on this rich dataset equips it to handle a wide range of scenarios, enhancing its robustness and generalization capabilities. This effectively enables \ours\ to forecast unseen time series accurately while eliminating the need for individual model training and optimization.

\subsection{Training \ours}

\ours\ underwent a multi-day training period on a cluster of NVIDIA A10G GPUs. During this process, we carried out extensive hyperparameter exploration to optimize learning rates, batch sizes, and other related parameters. We observed a pattern in alignment with findings from \citep{brown2020language}, where a larger batch size and a smaller learning rate proved beneficial. Implemented in PyTorch, \ours\ was trained using the Adam with a learning rate decay strategy that reduced the rate to 12\% of its initial value.

\subsection{Uncertainty quantification}

Probabilistic forecasting refers to estimating a model's uncertainty around the predictions. Correctly assessing a forecasting model's calibration enables risk assessment and informed decision-making. Conformal prediction, a non-parametric framework, offers a compelling approach to generating prediction intervals with a pre-specified level of coverage accuracy \citep{shafer2008tutorial, stankeviciute2021conformal}. Unlike traditional methods, conformal prediction does not require strict distributional assumptions, making it more flexible and agnostic to the model or time series domain. During the inference of a new time series, we perform rolling forecasts on the latest available data to estimate the model's errors in forecasting the particular target time series. 

\begin{figure*}[t]
\centering
\includegraphics[width=0.95\linewidth]{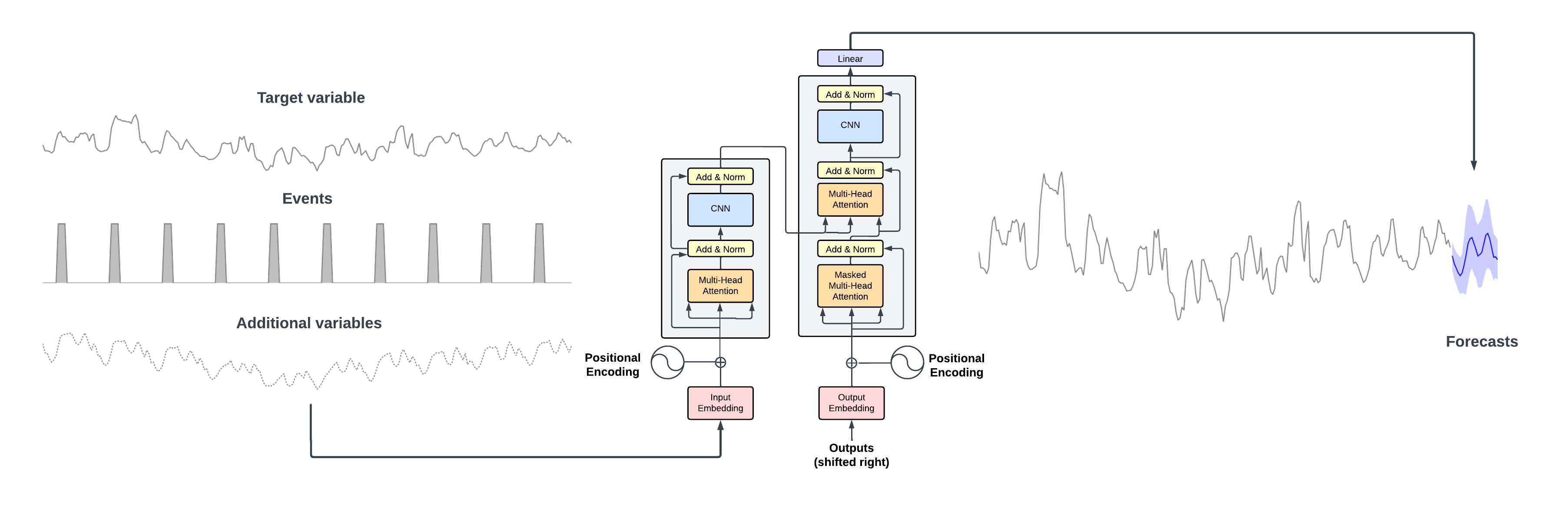}
\caption{Inference of new time series. \ours\ takes the historical values of the target values and additional exogenous variables as inputs to produce the forecasts. We rely on conformal predictions based on historic errors to estimate prediction intervals.}
\label{fig:forecasting}
\end{figure*}
\section{Experimental Results}\label{section:evaluation}

Classically, forecasting performance evaluation is based on splitting each time series of the dataset into train and test sets based on a defined cutoff. Such a principle, even in its cross-validation version, is not strict enough to asses a  foundation model because its main property is the capability to accurately predict completely novel series. 

In this section, we explore \ours's capabilities as a forecasting foundation model by testing it in a large and diverse set of time series that were never seen by the model during training. The test set includes over 300 thousand time series from multiple domains, including finance, web traffic, IoT, weather, demand, and electricity. 

The evaluation is performed in the last forecasting window of each time series, varying in length by the sampling frequency. \ours\ uses the previous historical values as inputs, as shown in Figure \ref{fig:forecasting}, without re-training its weights (zero-shot). We specify a different forecasting horizon based on the frequency to represent common practical applications: 12 for monthly, 1 for weekly, 7 for daily, and 24 for hourly data. \footnote{Future work would profit from expanding and varying this testing set.}

\ours\ was benchmarked against a broad spectrum of baseline, statistical, machine learning, and neural forecasting models to provide a comprehensive performance analysis. Baselines and statistical models are individually trained on each time series of the test set, utilizing the historical values preceding the last forecasting window. We opted for a global model approach for machine learning and deep learning methods for each frequency, leveraging all time series in the test set. Some popular models like Prophet \citep{taylor2018forecasting} and ARIMA were excluded from the analysis due to their prohibitive computational requirements and extensive training times.

\begin{figure*}[t] %
\centering
\includegraphics[width=0.8\linewidth]{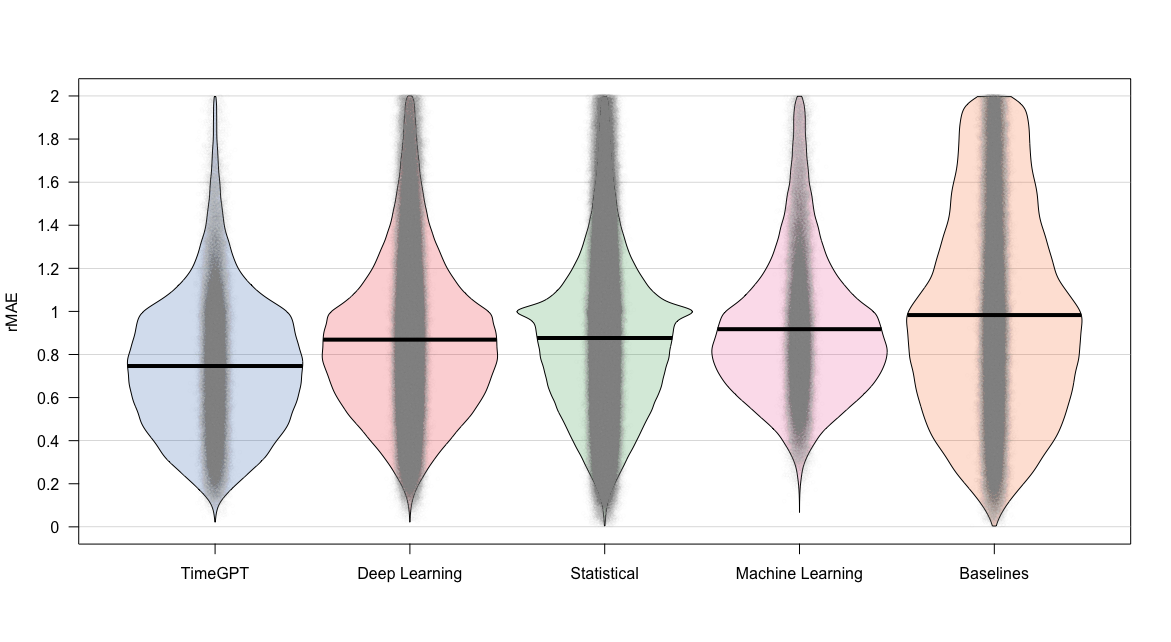}
\caption{Relative Mean Absolute Error (rMAE) for \ours\  and various groups of models for montly frequency. Each bean in the plot represents the rMAE distribution for a group, with the central line showing the mean. \ours\ leads in performance, followed by deep learning methods, statistical, machine learning, and baseline models. Results are analogous for other frequencies.}
\label{fig:boxplot}
\end{figure*}

Our selected evaluation metrics include the relative Mean Absolute Error (rMAE) and the relative Root Mean Square Error (rRMSE), both normalized against the performance of the Seasonal Naive model. This choice is justified by the additional insights offered by these relative errors, as they show performance gains in relation to a known baseline, improving the interpretability of our results. The relative error metrics bring the additional benefit of scale independence, enabling comparisons across the results for each frequency. To ensure both robust numerical stability and consistency in evaluation, we apply this normalization at a global scale for each comprehensive dataset. The specific computations for these metrics, applicable to a dataset with $n$ time series and a forecast horizon of $h$, are described in Equation \ref{eq:metrics}.

\begin{equation}\label{eq:metrics}
    rMAE = \frac{\sum_{i=1}^{n}\sum_{t=1}^{h} \left| y_{i,t} - \hat{y}_{i,t} \right|}{\sum_{i=1}^{n}\sum_{t=1}^{h} \left| y_{i,t} - \hat{y}^{base}_{i,t} \right|} \quad \quad rRMSE = \frac{\sum_{i=1}^{n} \sqrt{\sum_{t=1}^{h} \left( y_{i,t} - \hat{y}_{i,t} \right)^2}}{\sum_{i=1}^{n} \sqrt{\sum_{t=1}^{h} \left( y_{i,t} - \hat{y}^{base}_{i,t} \right)^2 }}
\end{equation}

\subsection{Zero-shot inference}

\begin{table*}[t]
\small
    \begin{center}
    \caption{Main performance results for \ours\ with zero-shot inference and benchmark models measured with rMAE and rRMSE, lower scores are better. The best model for each frequency and metric is highlighted in bold, the second best underlined, and the third best underlined with a dotted line.}
    \label{table:main_results}
    \setlength\tabcolsep{5.0pt}
	\begin{tabular}{l | cc | cc | cc | cc } \toprule
                         & \multicolumn{2}{c}{Monthly}     & \multicolumn{2}{c}{Weekly} & \multicolumn{2}{c}{Daily} & \multicolumn{2}{c}{Hourly}      \\
                         &     rMAE          &     rRMSE      &     rMAE      &     rRMSE      &     rMAE      &     rRMSE      &     rMAE      &     rRMSE   \\
    \midrule
        \ZeroModel       &     2.045         &     1.568     &     6.075     &     6.075     &     2.989     &     2.395     &     10.255    &     8.183 \\
        \HistoricAverage &     1.349         &     1.106     &     4.188     &     4.188     &     2.509     &     2.057     &     2.216     &     1.964 \\
        \SeasonalNaive   &     1.000         &     1.000     &     1.000     &     1.000     &     1.000     &     1.000     &     1.000     &     1.000 \\
    \midrule
        \ThetaModel      &     0.839         &     0.764     &     1.061     &     1.061     &     0.841     &     0.811     &     1.163     &     1.175 \\
        \DOTheta         &     0.799         &     0.734     &     1.056     &     1.056     &     0.837     &     0.806     &     1.157     &     1.169 \\
        \ETS             &     0.942         &     0.960     &     1.079     &     1.079     &     0.944     &     0.970     &     0.998     &     1.009 \\
        \CES             &     1.024         &     0.946     &     1.002     &     1.002     &     0.919     &     0.899     &     0.878     &     0.896 \\
        \ADIDA           &     0.852         &     0.769     &     1.364     &     1.364     &     0.908     &     0.868     &     2.307     &     2.207 \\
        \IMAPA           &     0.852         &     0.769     &     1.364     &     1.364     &     0.908     &     0.868     &     2.307     &     2.207 \\
        \CrostonClassic  &     0.989         &     0.857     &     1.805     &     1.805     &     0.995     &     0.933     &     2.157     &     2.043 \\
    \midrule
        \LGBM            &     1.050         &     0.913     &     0.993     &     0.993     &     2.506     &     2.054     &\textbf{0.733} &\textbf{0.709} \\
    \midrule
        \LSTM            &     0.836         &     0.778     &     1.002     &     1.002     &     0.852     &     0.832     &     0.974     &     0.955 \\
        \DeepAR          &     0.988         &     0.878     &     0.987     &     0.987     &     0.853     &     0.826     &     1.028     &     1.028 \\
        \TFT             & \dotuline{0.752}  &\dotuline{0.700}  &\dotuline{0.954} &\dotuline{0.954}&\dotuline{0.817}&\dotuline{0.791}&     1.120     &     1.112 \\
        \NHITS           & \underline{0.738} &\underline{0.694} &\underline{0.883}&\underline{0.883}&\textbf{0.788} &\textbf{0.771} &\underline{0.829}&\underline{0.860} \\
    \midrule
        \ours            &\textbf{0.727}     &\textbf{0.685}    &\textbf{0.878}   &\textbf{0.878} &\underline{0.804}&\underline{0.780}&\dotuline{0.852}&\dotuline{0.878} \\
    \bottomrule
	\end{tabular}
	\end{center}
\end{table*}

We first test \ours\ capabilities on zero-shot inference, meaning that no additional fine-tuning is performed on the test set. Table \ref{table:main_results} presents the zero-shot results. Remarkably, \ours\  outperforms a comprehensive collection of battle-tested statistical models and SoTA deep learning approaches, ranking among the top-3 performers across frequencies. 

It must be noted that the validity of a forecasting model can only be assessed relative to its performance against competing alternatives. Although accuracy is commonly seen as the only relevant metric, computational cost and implementation complexity are key factors for practical applications. In this regard, it is noteworthy that the reported results of \ours\ are the result of a simple and extremely fast invocation of the prediction method of a pre-trained model. In comparison, other models require a complete pipeline for training and then predicting.

\subsubsection{Comparison with recent Foundation Models}

Work in progress.

\subsection{Fine Tuning}

\begin{figure*}[t] %
\centering
\includegraphics[width=0.6\linewidth]{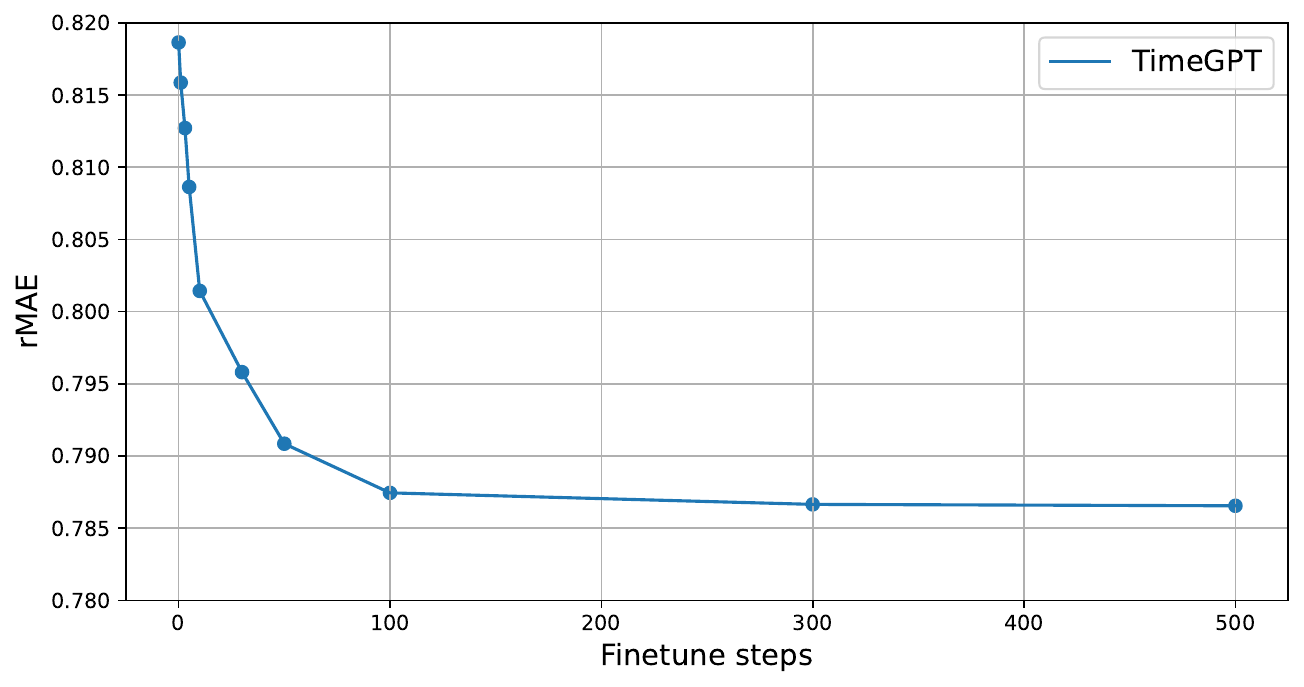}
\caption{\ours\ performance with fine-tuning on a subset of time series from the test set measured by rMAE.}. 
\label{fig:finetuning}
\end{figure*}

Fine-tuning is a critical step in effectively utilizing foundation models and transformer-based architectures. Foundation models are pre-trained on vast amounts of data, capturing wide-ranging and generic features. However, these models often need to be specialized for specific contexts or domains. By fine-tuning, we adjust the model parameters on a task-specific dataset, allowing the model to tailor its vast pre-existing knowledge toward the requirements of the new task. This process ensures that the model retains its broad understanding and excels at the specific tasks at hand. Due to their inherent flexibility and capacity for learning complex patterns, transformer-based architectures particularly benefit from fine-tuning, enhancing their performance in domain-specific applications. Fine-tuning thus serves as a crucial bridge, linking foundation models' broad capabilities to the target tasks' specificities. Figure \ref{fig:finetuning} presents results on the accuracy improvements of \ours\ against the number of fine-tuning steps for a subset of time series on the test set.

\subsection{Time Comparison}
For zero-shot inference, our internal tests recorded an average GPU inference speed of 0.6 milliseconds per series for \ours, which nearly mirrors that of the simple Seasonal Naive. As points of comparison, we consider parallel computing-optimized statistical methods, which, when complemented with Numba compiling, averaged a speed of 600 milliseconds per series for training and inference. On the other hand, global models such as LGBM, LSTM, and NHITS demonstrated a more prolonged average of 57 milliseconds per series, considering both training and inference. Due to its zero-shot capabilities, \ours\ outperforms traditional statistical methods and global models with total speed by orders of magnitude. 








\section{Discussion and Future Research} \label{section:discussion}

Current forecasting practice usually involves a complex pipeline, encompassing multiple steps from data processing to model training and selection. \ours\ greatly simplifies this process by reducing pipelines to the inference step, substantially reducing complexity and time investment while still achieving state-of-the-art performance. Perhaps most significantly, \ours\ democratizes the advantages of large transformers models, nowadays restricted to organizations with vast amounts of data, computational resources, and technical expertise. We believe that foundational models are set to profoundly impact the forecasting field and can redefine current practices.

The introduction of a foundation model in time series that resembles other fields and opens the possible path for future improvements could be considered an important milestone in the time series field. However, this work must be understood as part of a larger academic tradition with a plethora of open questions. While we believe that \ours\ displays amazing results presenting for the first time a general global modal capable of accurately predicting unseen series, there are still many important limitations and open questions. We hope this assessment is of help to current and future researchers.

Our results align with previous intuitions regarding the expected performance of large time series models. This is consistent with findings from Zalando, OpenAI, Alibaba, and Amazon \citep{kunz2023deep, brown2020language, eisenach2020mqtransformer}. These outcomes validate the scaling laws correlating model size, dataset size, and Transformer performance. These laws elucidate why simpler models might outperform Transformers on smaller datasets, as observed in studies such as \citep{zeng2023transformers}. The relevance of Transformers is, therefore, context-dependent, and they often become more beneficial as dataset sizes increase. These laws offer important practical insights, guiding model selection for specific tasks. In situations where there are limitations on the availability of large datasets or computational resources, simpler models might be more fitting.

Looking forward, we identify two primary areas for future exploration:

\begin{enumerate}
\item \textbf{Informed forecasting}: that incorporates knowledge about the underlying processes, such as physical laws, economic principles, or medical facts.
\item \textbf{Time Series Embedding}: While traditionally practitioners have hypothesized that series from the same categories like retail or finance would have greater similarity than those across domains, a robust metric to measure similarity between series could significantly benefit the field. This work suggests that certain assumptions around the taxonomy of time series warrant further examination.

\end{enumerate}

Furthermore, adjacent questions about foundation models for time series classification and the integration of truly multimodal (text, video) and multi-temporal foundation models promise to be engaging areas for future study. These areas will not only extend our understanding of time series data but also improve our ability to develop more powerful and generalized models for forecasting.

\section*{Acknowledgements} \label{section:acknowledgements}

We would like to thank Rob Hyndman, Tim Januschowski, Valeriy Manokhin, and Sean Taylor for their insightful comments that greatly improved our work. 
\clearpage
\bibliography{citations}
\bibliographystyle{plainnat}

\clearpage
\appendix
\section{Access and early testing}

\ours\ has undergone rigorous internal testing, demonstrating robust performance across various domains and frequencies. As we move forward, we invite practitioners and researchers to explore its capabilities on their own datasets and tasks. 

To facilitate this, we have released comprehensive guides that explain how to utilize \ours\ with features including uncertainty quantification, fine-tuning, forecasting multiple time series, integrating calendar and exogenous variables, and performing anomaly detection. 

\ours\ is accessible through both a \href{https://nixtla.github.io/nixtla/}{Python SDK} and a REST API endpoint, currently available in private beta. It has been designed with user-friendliness in mind, making it easy to implement even for those with limited experience in time series forecasting. \ours\ can be executed with just a few lines of code, streamlining the process for researchers and practitioners alike. The following snippet presents the pseudocode of a complete forecasting pipeline with \ours. \\

\begin{python}
from nixtla import NixtlaClient 

nixtla_client = NixtlaClient(api_key="YOUR API KEY HER")
fcst_df = nixtla_client.forecast(YOUR_DATA)
\end{python}

To request access, please visit \href{https://www.nixtla.io/}{nixtla.io}.

\end{document}